\documentclass[ acmsmall]{acmart}
\AtBeginDocument{%
  \providecommand\BibTeX{{%
    \normalfont B\kern-0.5em{\scshape i\kern-0.25em b}\kern-0.8em\TeX}}}

\setcopyright{acmcopyright}
\copyrightyear{2022}
\acmYear{2022}
\acmDOI{XXXXXXX.XXXXXXX}

 \acmConference[ACM SIGMETRICS 22']{ACM SIGMETRICS / IFIP PERFORMANCE 2022}{June 06--10,
 2022}{Mumbai, India}
\usepackage{algpseudocode}
\usepackage{algorithm}
\usepackage{subcaption}
\newtheorem{remark}{Remark}
\author[1]{Eliran Sherzer, Arik Senderovich, Opher Baron and Dmitry Krass}

\title{Can machines solve general queueing systems?}
\date{February 2022}

\begin{document}

\begin{abstract}
    Computing performance measures in \emph{general} queueing systems is a challenging task: most complex systems are intractable,
    while simpler systems
    often require involved mathematical analysis to derive exact solutions. For more complex systems only approximation formulas are available, which  may lead to large errors. While a common practical alternative is simulation modeling, a simulation model can only be designed for a specific, not a general system.  Moreover, such simulation models may take a long time to converge to an equilibrium solution.
    
    \noindent In this paper, we analyze how well a machine can solve a general problem in queueing theory. To answer this question, we use a \emph{deep learning} model to predict the stationary queue-length distribution of an $M/G/1$ queue (Poisson arrivals, general service times, one server). To the best of our knowledge, this is the first time a machine learning model is applied to  a general
    queueing theory problem. We chose $M/G/1$ queue for this paper because it lies ``on the cusp'' of the analytical frontier: on the one hand exact solution for this model is available, which is both computationally and mathematically complex. On the other hand, the problem (specifically the service time distribution) is general. This allows us to compare the accuracy and efficiency of the deep learning approach to the analytical solutions.
    
    \noindent The two key challenges in applying machine learning to this problem are (1) generating a diverse set of training examples that provide a good representation of a ``generic'' positive-valued distribution, and (2) representations of the continuous  distribution of service times as an input. 
    To deal with the first challenge we  propose a novel method, based on phase-type distributions, to sample service distributions such that our data will ``widely cover'' the space of non-negative distributions. We use the results from queuing theory to obtain  exact solutions of $M/PH/1$ models (phase-type service distribution). 
    
    \noindent To overcome the second challenge, we use the arrival rate and the first $n$ moments of the service times as inputs to our deep learning model. The output (that we learn) is the stationary queue-length distribution. To test our method, we conduct an extensive empirical evaluation. We use the sum of absolute errors between our predictions and the ``ground truth'' as our primary metric for evaluating the accuracy. 
    
    \noindent Our results show that our model is indeed able to predict the stationary behavior of the $M/G/1$ queue extremely accurately: the average value of our metric over the entire test set is $0.0009$; other metrics we employ also show an excellent predictive accuracy. Moreover, our machine learning model is very efficient, computing very accurate stationary distributions in a fraction of a second (an approach based on simulation modeling would take much longer to converge). We also present a case-study that mimics a real-life setting and shows that our approach is more robust and provides more accurate solutions compared to the existing methods.  This shows the promise of extending our approach beyond the analytically solvable systems (e.g., $G/G/1$ or $G/G/c$). 
    
    \noindent One interesting insight yielded by our approach is that the number of moments of the service time distribution required for accurately capturing the dynamics of an $M/G/1$ queue is quite small; high accuracy is achieved with $n=5$, while the higher service time moments have negligible effect on the accuracy.
\end{abstract}

\maketitle


\section{Introduction} \label{sec:intro}

In this paper we attempt to ``teach'' a machine to compute performance measures in a general queueing system. In effect, we ask whether a machine can replace complicated mathematical methods 
and provide an accurate solution for problems in queueing theory. As we show in this paper the answer is likely in
the \emph{affirmative}.  

In order to investigate how well a machine can solve a general problem from the queueing domain, we 
revisit the computation of the stationary distribution of an $M/G/1$ (Poisson arrivals, general service, single server) queue, 
a well-studied problem in queueing theory, and solve it using machine learning (ML). We focus
on this problem, because it lies ``on the edge'' of the analytical frontier: while  exact solutions for this model are available, they are both computationally and mathematically complex. On the other hand, the problem (specifically the service time distribution) is general. This allows us to compare the accuracy and efficiency of the deep learning approach to the analytical solutions. 

\noindent \textbf{Motivation} Typically, queueing systems are complex, hard to analyze, and analytical methods often struggle even with simple stylized models. For example, a 
single-server queue with 
general inter-arrival and service distributions (the $G/G/1$ queue) 
cannot be analyzed in an exact fashion, 
and robust approximation methods are not available. 
Furthermore, real-life queueing systems introduce complexities that 
often make queueing models analytically  
intractable, e.g.,  customer abandonment, correlations between arrivals and service times, and many more. Thus, in many cases, since no  analytic solution is available, the system can only be analyzed via simulation. 
Yet, simulation has
its own downsides, ranging from the requirement for a detailed specification of all underlying distributions to, often,  a 
long convergence time, especially for multi-server systems. 

In this work, instead of using the traditional queueing theory methods (e.g., embedded Markov-chains at departure epochs or supplementary variables),
we leverage the power of ML, using deep neural networks as our solution framework. 
This is motivated by the major successes of this approach in a variety of fields ranging from computer vision, to natural language processing, to traffic prediction in transportation systems (see \cite{LeCun2015,726791, 5537907}). Moreover, within the Operations Research (OR) domain, recent papers have shown the usefulness of deep learning in well-known problems such as scheduling, resource allocation, and others (see~\cite{222611, 8943940}). 

\noindent \textbf{Technical challenges} Applying a deep neural model to solve\footnote{Throughout this paper, ``solve'' refers to computing the stationary queue-length distribution} an $M/G/1$ system is a non-trivial task. One faces several major challenges:
 \begin{enumerate} \item Generating a set of training
instances that provide a good representation of a ``generic'' positive-valued distribution. More specifically, we need to obtain a sufficiently diverse set of input pairs of arrival rates and service-time distributions that would be representative of the space of all non-negative distributions, while maintaining a ``conformable'' parametric representation.
 \item Representation
of service time distributions as an input to the deep learning model. While traditional 
steady-state computation methods use arrival rates and service time distributions as direct inputs, 
an input to a deep learning model must be represented by a tensor; with possibly some information being lost as a result. What is the right distribution-to-tensor mapping? 
\end{enumerate}

\noindent To deal with the first challenge, we leverage several key results for Phase-Type (PH) family of distributions. The power of PH distributions lies in the fact that they are dense in the class of all 
non-negative
distribution functions. Practically speaking, this means that a PH distribution
with a sufficient number of phases can, theoretically, approximate any non-negative distribution
to an arbitrary precision (see~\cite{Asmussen2003}, Theorem 4.2).    Finally, an $M/PH/1$ system (PH-type service time distribution) allows for a much more computationally efficient solution methodology than a general $M/G/1$ queue.  
Sampling from the universe of PH distributions
is not a straightforward task, as its parameters are not independent and have multiple constraints that must be enforced. 
To overcome this challenge, we propose a novel and versatile 
sampling method that covers a ``wide'' range of distributions. We show the latter claim empirically in Section~\ref{sec:data}.  

\noindent To address the second challenge, we develop a representation of the service time component of the input as the first $n$ moments of the service time distribution; these moments are computed from the empirical data set of observed service times. The number of moments $n$ is a hyper-parameter of our 
approach, which is determined after examining the accuracy of the learned model as function of $n$.  

\noindent While representing the service time distribution using a finite number of moments may seem counter-intuitive (since different distributions can have identical set of first $n$ moments) this representation is natural in view of a number of existing results in stationary queueing analysis which often depend only on the  first two moments of both the service time and the inter-arrival time distributions; for example, the mean queue length for the $M/G/1$ queue given by the Khintchine–Pollaczek formula, or the bound of the mean queue length in the $G/G/1$ queue (see Chapter 2 in~\cite{https://doi.org/10.1002/net.3230070308}). The first two moments are also widely used in more sophisticated queueing models such as queueing networks (see~\cite{Whitt1994, https://doi.org/10.1111/j.1937-5956.1993.tb00094.x}). 
While it is clear that the moments of the  distribution play a key role in queueing analysis, we do not know how many moments are actually needed to have a sufficient representation of 
the entire queue length distribution. This study allows us to shed some light on this important question. 


\noindent \textbf{Results} Our results show that a deep neural network can almost perfectly approximate the stationary queue-length distribution for $M/G/1$ queues, while
using a relatively small net (roughly 15,000 parameters). Hence, potentially, there is room for using deep learning techniques for more complex queueing systems. Further, we show that the only the first $5$ moments of service time suffice to determine the distribution of the $M/G/1$ stationary queue length; there is no  substantial gain in accuracy from adding higher-order service time moments. This result is very intriguing as it provides further understanding of the dynamics of an $M/G/1$ queue, and perhaps more complex systems as well. It also greatly simplifies the task of fitting a particular parametric distribution to the observed service time data as only the first $5$ moments have to be matched, rather than using more complex methods such as EM (Expectation Maximization algorithm)\footnote{Papers like~\cite{TELEK20071153} are doing just that.}. 

\noindent Note that by applying deep learning, we are able to provide a new data-driven solution
to a fundamental problem in queueing theory. To use the traditional analytical $M/G/1$ results in practice, when one is provided with data, rather than a closed-form distribution of service times, one must carefully choose a parametric family of distributions, apply a statistical or an algorithmic method in order to find the parameters that optimize some notion of goodness-of-fit, and only then apply the analytical technique to obtain steady-state queue-length distribution. These procedures are seldom straightforward, are often time consuming, and do not guarantee an accurate fit in the end.  Our approach, where only the first $5$ moments are computed and then the solution is directly produced by a pre-trained neural network is far more simple, computationally efficient, and accurate, as demonstrated in Section~\ref{sec:casy_study}. 

\noindent \textbf{Contributions} The main contributions of this paper are threefold:
\begin{itemize}
    \item We provide a novel data-driven technique based on deep learning to solve
    a fundamental problem in queueing theory. The network takes as input a compact representation of the problem by using only the arrival rate and the first $n$ moments of the service time distribution.
    \item We introduce a new sampling algorithm for PH service time distributions and show how it can be used to generate a set of training samples. 
    The algorithm guarantees various important properties for the sampled PH distribution, including absorption and stability. 
    \item We provide an extensive empirical evaluation of the approach by conducting a set of experiments as
    well as analyzing a case-study that mimics a real-world scenario. 
\end{itemize} 

\noindent \textbf{Organization} The remainder of the paper is organized as follows. 
In Section~\ref{sec:lit_rev} we present an overview of the related literature.
Section~\ref{sec:overview} provides a high-level end-to-end overview of our solution.
We provide our PH sampling technique in Section~\ref{sec:data},
describe our neural architecture, input pre-processing, and loss function in Section~\ref{sec:deep}. 
The main findings from our empirical evaluation are provided in Section~\ref{sec:evaluation},
while a realistic case-study is provided in Section~\ref{sec:casy_study}. We conclude the paper
with a discussion of the limitations of our approach (Section~\ref{sec:discussion}), and
some concluding remarks and directions for future work in Section~\ref{sec:conclusions}.

\section{Related Work}\label{sec:lit_rev}

\subsection{Computing the Stationary Queue-Length Distribution}

There are two well-known analytic methods for computing the stationary queue length of the $M/G/1$ queue. One is via the probability generating function  given by the Pollaczek–Khinchine transform equation (see~\cite{Harchol2013}). The other is via the supplementary variable method by Cox and Kendall~\cite{cox_1955}. While both methods are exact, they can be computationally burdensome: in order to obtain the probability of having $n$ customers in the steady-state system, the former method requires the $n^{th}$ derivative to the generating function, while for the latter we need to solve a set of $n$ non-linear equations. Even if service times are uniformly distributed both methods can be computationally heavy. 

Since our training sample is obtained using PH distributions, a simpler option is available for computing the stationary queue-length distribution: the Quasi Birth-and-Death method (QBD) developed by Neuts~\cite{Neuts1981}. As mentioned in Section~\ref{sec:intro}, a PH distribution
with a sufficient number of phases can approximate any non-negative distribution
arbitrarily closely. However, as we show later in the paper, it is not always easy to fit a PH representation that 
provides a good approximation for a given data trace. 

\subsection{Deep Learning for Queueing Systems}
There have been only a few prior  studies that utilize deep learning neural networks to analyze queues.  The most similar paper to ours is by Nii {\em et. al.}~\cite{Nii20} where the authors propose a deep learning model to analyze a $GI/G/s$ queue.  The input includes only the first two moments of both the inter-arrival and the service times distribution. However, they are evaluating only the average waiting time, rather than estimating the whole distribution, and consider only a small range of possible values using a very small NN (roughly 100 variable and three hidden layers). Kyritsis and Deriaz~\cite{Kyritsis19} also utilized a neural network to predict waiting times. However, unlike our model or the method proposed in~\cite{Nii20}, they do not refer to a steady-state distribution, but rather focus on an online prediction setting conditioned on the current state of a network. Their results are not sufficiently robust to make predictions on new data sets. In a similar work to~\cite{Kyritsis19},  Olawoyin and Hijry~\cite{Hijry_2020}  evaluate the expected waiting times in an emergency room using deep learning. This is quite different from our goal of studying the ability of a machine learning approach to analyze a general queuing problem. 

Other papers that combined queueing systems and machine-learning did not employ deep learning for evaluating queueing systems, focusing instead on optimizing and controlling the queue. Efrosinin and Stepanova~\cite{Efrosinin} used deep learning for estimating of the optimal threshold policy in queues with
heterogeneous servers. Dai and Gluzman~\cite{dai2021queueing} used a deep reinforcement learning model for controling queueing networks by effectively solving a Markov decision process. For additional papers that employ machine-learning for optimization and control of queueing networks we refer the reader to~\cite{CHEN20113323, 8919665, 9502440}. These papers utilize machine learning methods to solve an optimization problem, while we are using them for capturing the stochastic dynamics of the system. 

\section{Solution Overview}
\label{sec:overview}
\begin{figure}
\centering
\includegraphics[scale=0.55]{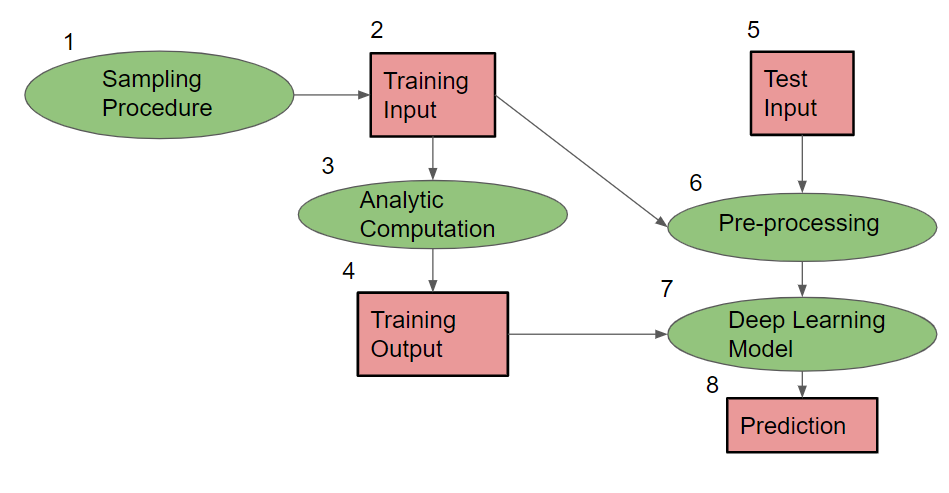}
\caption{Work-flow diagram of our learning procedure. }
\label{fig:diagram}
\end{figure}

The main phases 
of our solution are presented in~Figure~\ref{fig:diagram}, where green ovals correspond to processing/computational steps and red rectangles to data sets. 

Steps 1-4 represent the generation of training input and output data; they are described in Section \ref{sec:data} below. We start by generating input samples (step 1, ``Sampling Procedure'' on Figure~\ref{fig:diagram}); see Section \ref{sec:sampling} for details. The sample is generated in the form of pairs, with each pair comprising an arrival rate and a service distribution\footnote{we only consider pairs that result in a stable system, i.e., the arrival rate must be smaller than the service rate.}.  This results in the ``Training Input'' data set in (step 2 on the Figure).
Next we compute the stationary queue-length distribution  by applying a well-known QBD method   (step 3, ``Analytic Computation'' on the Figure; details can be found in Section \ref{sec:output}). 
This results in the training output (step 4), used as the target when learning from the training data. The output consists of the first $l$ values of the stationary queue-length distribution; the value of $l$ is described below.

Steps 2, 6, 7 and 8 on Figure~\ref{fig:diagram} correspond to the training of the Deep Learning model, and steps 5,6,7,8 to application of the model to new test data. These steps are detailed in Section \ref{sec:deep}. 

Prior to entering the input into the neural network some pre-processing is required (step 6, ``Pre-processing'' in Figure~\ref{fig:diagram}). First, we compute the first $n$ moments of the service time distribution, where the value of $n$ is a hyper-parameter that requires tuning.  Next, the computed moments undergo some additional pre-processing (standardization).  These steps are detailed in Section \ref{sec:intput preprocessing}.
The processed input and the training output are fed into the deep learning model for training (step 7,  ``Deep learning Model'' on Figure~\ref{fig:diagram}); see Section \ref{sec:network} for details. The model outputs the predicted stationary queue-length distribution (step 8) in the same format as the training output.

Once the training is completes, we use test data (step 5), apply the required pre-processing, and use our deep learning model to obtain the predicted values. 



\section{Data Generation}\label{sec:data}

In this section we outline the input sampling procedure (Section~\ref{sec:sampling}), and the training output generation step (Section~\ref{sec:output}). 


\subsection{Input Sampling Procedure}\label{sec:sampling}

To ensure the stability of the queueing system we must make sure the arrival rate is smaller than the corresponding service rate. To achieve this, it is simpler to sample the service time distribution first, and subsequently obtain the arrival rate. The latter is
drawn uniformly over the stability range of the sampled service rate. The main challenge here is to sample many service distributions so that they provide a ``wide'' coverage of the state-space of all positive distributions. To this end, we employ phase-type (PH) distributions; as mentioned earlier, PH distribution can approximate any non-negative distribution~\cite{Neuts2013}.

\noindent \textbf{Phase-Type Distribution} A PH distribution can be represented by a random variable describing the 
time until absorption of a continuous-time Markov chain (CTMC) with a single absorbing state; all other states are transient.
Each of the transient states of the CTMC corresponds to 
one of the states of the PH distribution, where a state sojourn time follows an exponential distribution. For a full representation of PH with $m\geq 1$ states, 
we need to specify two parameters, namely $\alpha$, the initial probability $m$-vector of starting the process in any of the $m$ states, and $S$, the corresponding generating $m\times m$ matrix of the CTMC. We restrict our method to PH distributed random variables such that $\sum_{i}|\alpha_i|=1$\footnote{Implying that we do not allow the system to start in the absorbing state.}. In order for the pair $(\alpha,S)$ to be a valid pair for the CTMC, we need to meet the following constraints:
\begin{itemize}
    \item $S[i,i]<0, \forall i \leq m$.
    \item $\sum_{j=1}^{m} S[i,j] \leq 0$. 
\end{itemize}

\noindent \textbf{Challenges in Sampling from PH} For a general input, we wish to control two degrees of freedom hidden in $\alpha$ and $S$: (1) the size (number of states) of the PH distribution, and (2) the transition function that governs the switches between the different states (to be explained below). The former is relatively straightforward to deal with as we uniformly sample a value between $1$ and $max_{ph}$, where $max_{ph}$ is the 
maximum PH size that we use for our experiments (this is a hyper-parameter of our approach and can be tuned by the user). 
The second consideration is not trivial and requires some effort. 

When generating $\alpha$ and $S$ we wish to allow all possible transition structures from state to state. The transition structure is determined by three elements: (i) the initial probabilities vector $\alpha$, (ii) the locations of the non-diagonal cells in $S$ in which the value is $0$, and (iii) the sum of each row in $S$ . Note that iff $S[i,j]=0$, for $i \neq j$, then it is impossible to directly reach state $j$ from state $i$. Clearly, if all non-diagonal cells in a single row $i\leq m$ are $0$, then the underlying CTMC will reach its absorption state once the system exits state $i$. We refer to such states as \textit{fully-absorbing}. These are controlled by element (ii).  Letting $s_i$ be the sum of row $i$ in $S$, i.e., $\sum_{j=1}^{m}S[i,j]=s_i$, if $s_i<0$, then with probability $s_i/S[i,i]$  the CTMC will reach its absorption state after exiting state $i$.  We refer to such states as \textit{partially absorbing}. However, if $s_i=0$, then the CTMC cannot reach the absorption state once the system exits state $i$. We refer to such states as \textit{non-absorbing}. The composition  of partially absorbing and non-absorbing states is controlled by element (iii). Thus, in order to control the structure of the PH distribution we need to control for elements (i), (ii), and (iii).

Note that we can have different values of $\alpha$ and $S$ and yet still end up having the same service time distribution. 
Moreover, in queueing systems, 
the time scale does not matter. 
What matters is only the ratio between 
inter-arrival and service times. Thus, when having 
two different distributions, they may still be very similar once we put the two 
on the same time scale. In order to avoid such cases and sample distributions as diverse as possible, 
we first split states into ``classes''. 

\noindent \textbf{Classes of States in PH} We define a \textit{class} as a set of states such that if we reach one of them from the initial state, it is impossible
to reach any state outside this set before the underlying CTMC reaches its absorption state. Each class, in fact, is a PH distribution on its own. Thus, we can sample each class separately and then group them together for a singe PH representation. 
The main reason for doing so is that if we have a single large class, 
even with different structures of transition matrices, we will end up in many cases with similar distributions~\cite{Neuts2013}. 
Since class structures are crucial for the generality of our result we shall first sample the 
number of classes and their sizes and only then sample the specific states within each class. 

The total number of classes may vary from 1 to $m$ as it cannot exceed the total number of states. We note that the PH random variable
(that corresponds to the time until the CTMC is absorbed) has, by its definition, a finite expected value. In order to avoid infinite expected values, each class must adhere to two conditions: (1) having at least a single partially or fully absorbing state, and (2) any  non-absorbing state must have a positive probability to reach one of the partially or fully absorbing states. Otherwise, the system may enter a ``livelock'', i.e., a loop in which it never reaches the absorbing state. We enforce condition (1) in our sampling methodology; however, enforcing condition (2) may increase the computational complexity of our sampling approach. Instead, once the procedure is complete, we check for finite expected value. If it the value is infinite, we reject the sample. We note that these rejections occurred in our extensive experiment only 5\% of the times (see Section~\ref{sec:evaluation} for details on the experiment).


\begin{algorithm}[t]
  \caption{Sample service time distribution}\label{algo:main}
   \begin{flushleft}
        \textbf{INPUT:} $max_{ph}$\\
        \textbf{OUTPUT:} $\alpha$, $S$ 
\end{flushleft}
   
  \begin{algorithmic}[1]
    \Procedure{Sampling PH}{}
      \State $m \gets U(1,max_{ph})$ 
      \State $NumClasses \gets U(1,m)$ \Comment{Sample number of different classes. }
      \State $P \gets Dirichlet(a_1,a_2,\ldots,a_{NumClasses})$ where $a_j \gets U(0,1), \forall j \leq NumClasses $
      \State $AssignedStates \gets 0$ 
    \For{$i=1,i\leq NumClasses,i++$ }
      
        \State $Size_i \gets U(1,m-AssignedStates-(NumClasses-i))$ 
        \State $AssignedStates \gets AssignedStates + Size_i$ 
        \State $\alpha^i, S^i \gets$ Sampling Class $i$ ($Size_i$) \Comment{Sample PH distribution for each class}
        \State $i_{low} = 1+\sum_{j=1}^{i-1}Size_j$ \Comment{$i_{low}$ is the lower index of the $i^{th} $ class in $\alpha$ and $S$}
        \State $i_{up} = 1+\sum_{j=1}^{i}Size_j$ \Comment{$i_{up}$ is the upper index of the $i^{th} $ class in $\alpha$ and $S$}
        \State $S[i_{low}:i_{up},i_{low}:i_{up}] = S^i$ 
        \State  $\alpha[i_{low}:i_{up}] = \alpha^i P[i]$ 
        \EndFor
        \If {$E[X]<\infty$, where $X \sim PH(\alpha,S)$}    
     \State Return $\alpha$, $S$
     \EndIf
    \EndProcedure
  \end{algorithmic}
\end{algorithm}


\begin{algorithm}[t]
  \caption{Sample PH of a single class}\label{algo:sample_class_i}
       \begin{flushleft}
        \textbf{INPUT:} $Size_i$\\
        \textbf{OUTPUT:} $\alpha^i$, $S^i$ 
\end{flushleft}

  \begin{algorithmic}[1]
    \Procedure{Sampling class $i$($Size_i$)}{}
      \State$\alpha^i \gets Zeros(1,Size_i), S^i \gets Zeros(Size_i,Size_i)$, $Trans \gets Zeros(Size_i,Size_i)$\
      \State $FullA_i, PartA_i, NonA_i \gets $ Sampling state types $(Size_i)$ \Comment{Assigning states to different types}
      \State{$Trans \gets$ Sample Trans($Trans, Size_i,  PartA_i, NonA_i$) }
      \State $S^i[j,j] \gets -U(1,1000), \forall j  \{j \leq Size_i\}  $
      \State $S^i[j,k] \gets U(1,1000), \forall j,k \{j\neq k,Trans[j,k]=1\}  $ 
     \State $p_j \gets U(0,1)$, $\forall j \in PartA_i $  \Comment{$p_j$ is the absorbing probability of state  $j$}

  \State{$S^i[j,k] = S^i[j,k] (1-p_j) \frac{-S^i[j,j]} {\sum_{l=1,l\neq j}^{Size_i}S^i[j,l]}$} ,$\forall j\in PartA_i , \forall k\neq j$
  \State{$S^i[j,k] = S^i[j,k] \frac{-S^i[j,j]} {\sum_{l=1,j\neq j}^{Size_i}S^i[j,l]}$}, $\forall j\in PartA_i \forall, k\neq j$

\State $\alpha^i \gets Dirichlet(a_1,\ldots,a_{Size_i})$, $a_j  \sim U(0,1), \forall j \leq Size_i$
\State Return $\alpha^i$, $S^i$
    \EndProcedure
\Procedure{Sample Trans($Trans, Size_i,  PartA_i, NonA_i$)}{}
\State $NumTrans_j \gets 1+Binomial(Size_i-2, U(0,1))$, $\forall j \in PartA_i, NonA_i$  
\State $Seleced_j \gets Choose(\{k\leq Size_i, k \neq j\},NumTrans_j)$ , $\forall j \leq Size_j$
\State $Trans[j,k]=1,$ $\forall j\leq Size_i, k \in Seleced_j $
\State Return $Trans$
\EndProcedure
  \end{algorithmic}
\end{algorithm}


\begin{algorithm}[t]
  \caption{Assign states into different classes}\label{algo:sample_FullA_partA_null_A}
    \begin{flushleft}
        \textbf{INPUT:} $max_{ph}$\\
        \textbf{OUTPUT:} $FullA_i$, $PartA_i$, $NonA_i$  
\end{flushleft}
   
  \begin{algorithmic}[1]
    \Procedure{Sampling state types($Size_i$)}{}
      \State{$FullA_i \gets \{\}$, $PartA_i \gets \{\}$, $NonA_i \gets \{\}$} 
      \State $p_1, p_2, p_3 \gets Dirichlet(a_1,a_2,a_3)$, where $a_j \gets U(0,1), j=1,2,3$
      \For{$j=1, j\leq Size_i, j++$} 
      \If{$j=1$} \Comment{To ensure at least one state in $FullA_i$ or $PArtA_i$} 
      \State{$state \gets Categorical(p_1/(p_1+p_2),p_2/(p_1+p_2))$} 
      \Else
      \State{$state \gets Categorical(p_1,p_2,p_3)$}
      \EndIf
      \State Assign state to $FullA_i$ if $state =1 $, $PartA_i$ if $state =2 $, $NonA_i$ if $state =3 $,
      \EndFor
     \State Return $FullA_i$, $PartA_i$, $NonA_i$
    \EndProcedure
  \end{algorithmic}
\end{algorithm}

\noindent \textbf{Sampling Algorithms} The sampling procedure is described in our main algorithm, Algorithm~\ref{algo:main}, while  Algorithms~\ref{algo:sample_class_i} and~\ref{algo:sample_FullA_partA_null_A} are procedures called from within Algorithm~\ref{algo:main}.

The idea of Algorithm~\ref{algo:main} is partitioning the PH into classes, sample each class independently (which takes place in Algorithm~\ref{algo:sample_class_i}) and then aggregate all classes into a unified PH representation $(\alpha, S)$.  We first sample the number of states $m$\footnote{Choosing $1000$ was a design choice. In principle, this parameter can be chosen by the user.} of the PH, the number of classes ($NumClasses$)  and the initial probability  of each class ($P$). Once those are obtained, we sample the PH representation of class $i$ (i.e., $\alpha^i$ and $S^i$), $\forall i\leq NumClasses$, given its size $(Size_i)$. This sampling procedure takes place in Sampling Class $i$ (i.e., the procedure in Algorithm~\ref{algo:sample_class_i}). We note that sampling $Size_i$, should be done carefully, as the constraint $\sum_{i=1}^{NumClasses}Size_i=m$ must be met. For this purpose, $Size_i$ is  sampled uniformly  over 1 to $m-AssignedStates-(NumClasses-i)$. The upper-bound ensures that we assign only from the remaining states, (i.e., that are not assigned by the previous classes: $1,\ldots,i-1$), and that there will be at least one state for all future classes (i.e,. states $i+1,\ldots,NumClasses$).

Once $\alpha^i$ and $S^i$ are sampled, we assign their values according to their designated locations in $\alpha$ and $S$, as described in lines 12 and 13. We note that for $S$ the assignment are straightforward, while in $\alpha$ we multiply them by $P[i]$. This is due to the fact that ``Sampling class $i$'' returns $\alpha^i$, which is summed up to $1$. However, the total probability of going into class $i$ is $P[i]$ and hence $\alpha^i$ is multiplied by $P[i]$. Finally, as mentioned above it is possible that the underlying CTMC will reach a livelock, hence we return $\alpha$ and $S$ only if the expected value of the sampled service time is finite.

\begin{remark}
Note that we sample from the Dirichlet distribution with random weights  throughout our sampling procedure (see also Algorithm~\ref{algo:sample_class_i} and~\ref{algo:sample_FullA_partA_null_A}). This is a useful way to sample a vector of probabilities (i.e., a vector that sums to 1) since it allows us to  get unbalanced,  as well as balanced, realizations. For example, if we sample $(a_1,a_2,a_3)$ from $U(0,1)$ and then normalize the weights to sum to 1, it is very unlikely to get an unbalanced realization like $(0.98,0.01,0.01)$. However, we want realizations that are as diverse as possible. When using a Dirichlet distribution, realization such as these are not that scarce. 
\end{remark}

Next, we provide details of the two sampling procedures used by Algorithm~\ref{algo:main}, first with Algorithms ~\ref{algo:sample_class_i} and then~\ref{algo:sample_FullA_partA_null_A}. Algorithm~\ref{algo:sample_class_i}  samples the $i^{th}$ class in our CTMC. The logic of the sampling procedures is as follows. We first classify each state and then determine all possible direct transitions. Naturally, the next step should sampling the rates where the transitions are possible. However, within any row of $S^i$, the transition rates are not independent. For example, if state $j$ is non-absorbing, then the sum of the $j^{th}$ row must be 0. In order to  avoid the  complexity of the conditional sampling we  take a different approach, which is first sample without conditioning and then normalize the rates accordingly such that all constraints are met. Once $S^i$ is determined we sample $\alpha^i$, which is done according to Dirichlet distribution.

For the sake of clarity, we would like to point out that each state $j\leq Size_i$ is classified to be either full-absorbing, partial-absorbing and non-absorbing and assigned to  $FullA_i$, $PartA_i$ and $NonA_i$, respectively.  $Trans$ is a $(Size_i,Size_i)$ matrix where $Trans[j,k]=1$ only if a direct transition from $j$ to $k$ is possible. $Trans$ is sampled in `Sample Trans', where the number of \textbf{potential} states that can be reached directly from state $j$  are $Size_i-1$. However, if $j \in PartA_i, NonA_i$ then there has to be at least one state $k\neq j$ to transfer to. This is taking care of in line 14 of Algorithm 2. Once $Trans$ is fixed, we sample  both diagonal and non-diagonal states, and then normalize the rates such that they will meet the constraints according to their type. For full-absorbing states nothing is requires as there is only one non-zero value, which is on the diagonal.  For normalizing
the rates of partial-absorbing and non-absorbing states, the  idea is similar in both cases: $S^i[j,k]$ is being multiplied by a constant $\frac{-S^i[j,j]} {\sum_{j=1,j\neq k}^{Size_i}S^i[j,k]}$, which makes sure that the sum of all non-diagonal values will sum to the diagonal value. For partial-absorbing states we also multiply by $1-p_j$, whereas, $p_j$ is the probability that state $j$ will reach the absorbing state upon departure.  This is because, we wish that the ratio between the sum of non-diagonal entries and the absolute value of the diagonal entry will be the probability that the process does not reach its absorbing state once it departs state $j$.

We next provide details of Algorithm~\ref{algo:sample_FullA_partA_null_A}. Here, we split all states $(1,\ldots,Size_i)$ into $FullA_i$, $PartA_i$ and $NonA_i$, which represent the full-absorbing, partial-absorbing and non-absorbing sets, respectively. It is important that at least  one of the sets $FullA_i$ and $PartA_i$ will be non-empty (otherwise, it is impossible to reach the absorbing state). We commence by sampling the probabilities $p_1$, $p_2$ and $p_3$, which represent the probability that a state assigned to $FullA_i$, $PartA_i$ and $NonA_i$, respectively (line 3). Then we iterate over all states, where the first state must be in either $FullA_i$ or $PartA_i$. This is done using the Categorical distribution. That is, a value is $i$ with probability $p_i$, for $i=1,2,3$. In line 6, the probabilities of the categorical distribution are normalized to sum up to 1.  Once the variable $state$ is sampled, we assign state $j$ to its designated set in line 10.

\begin{figure}
\centering
\includegraphics[scale=0.6]{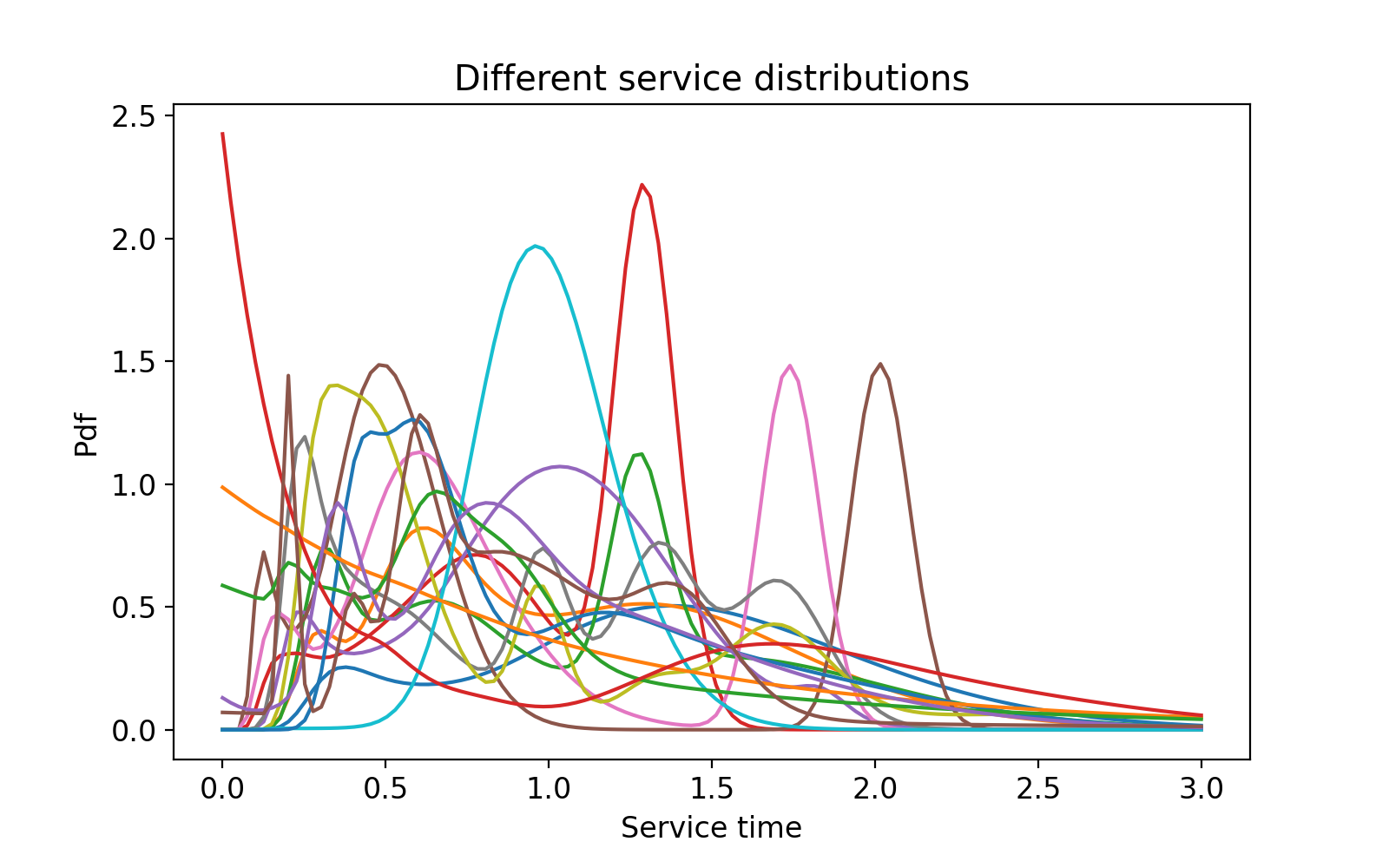}
\caption{An example of 15 service time  distributions sampled using Algorithm~\ref{algo:main}.  }
\label{fig:pdf_dist}
\end{figure}

\noindent \textbf{Illustration of the Results.} We empirically show that provided enough samples, 
our approach produces a set of phase-type distributions that provides a ``wide cover'' of the positive-valued distributions. For the sake of illustration, and without the loss of generality, we 
scale all service time distributions such that the mean value is 1. As shown in Figure~\ref{fig:pdf_dist}, the PH distributions generated by the Algorithms 1-3 above can take on various forms and capture complex structures. 

For further empirical evidence, we generated a data set containing 60,000 samples, which was later used as our test set. The mean value was set to $1$. We present the range of the service times for 6 different percentiles (see Table~\ref{table:precetiles_support}). The purpose  of this table is to demonstrate that we are indeed generating a very diverse set of distributions with  our sampling method. For each percentile level we provide  the minimum and the maximum service time in our data set. For example the ``Lower'' value of $8.511e^{-6}$  and the ``Upper'' value of $1.0715$ in the $25\%$ percentile column shows the very wide range of service times corresponding to this percentile in our sample of 60,000 service time distributions. Since we know that all distributions have a mean value of 1, the very low and high values shows a large variability in where the probability mass is concentrated among the distributions in our sample. This indicates a wide range of the types of distributions included in our sample, which was the goal of our sample generating process.  

\begin{table}[t]
\small
\centering
\begin{tabular}{||c |c|c|c|c|c|c||} 
 \hline
 Percentile & 25\%& 50\%& 75\% & 90\%& 99\%& 99.9 \%  \\ [0.5ex]
 \hline\hline

Lower &  $8.5111e^{-6}$ & 0.0003 & 0.0027 & 0.0308  & 0.8391 & 1.1684 \\ [0.5ex] 
 \hline\hline

Upper  & 1.0715 & 1.5812 & 3.0376 & 7.7300 & 29.9446 & 29.9871  \\ [0.5ex] \hline

\hline
\end{tabular}
\caption{Range of  percentiles.}
\label{table:precetiles_support}
\vspace{-1.5em}
\end{table}

\noindent Once the service time distribution (a PH) is sampled, we scale it such that its expected value is $1$.
To conclude input generation procedure, we sample arrival rates from $U(0,\rho_{max})$ (in our experiments, we chose $\rho_{max}=0.95$ to ensure that we do not cover very heavily loaded systems). Note that our scaling of mean service times to $1$ and arrival rates to $(0, 0.95)$ does not result in any loss of generality with respect to the steady state probabilities for systems with the utilization ratio in this range.

\begin{remark}
High utilization values, e.g., larger than 0.95, change the queue dynamics significantly, increasing the probability of very high queue lengths. If prediction for systems with very high utilization rates  is desired, the model should be trained only on samples corresponding to very high utilization values of the queue.  
\end{remark}

\subsection{Generating Output}\label{sec:output}

To generate the output (stationary queue-length distribution), we use  the arrival rate and PH representation of service time distribution (per input) to compute the first $l$ steady-state probabilities using the QBD method~\cite{Neuts1981}, which is significantly faster than the other analytical methods for the case of PH-distributed service times. 

Since we need to feed the computed distribution to the deep learning model in the form of a fixed vector, we truncate our computation at $l$ such that the total probability of having more than $l$ customers is smaller then $\epsilon$ (these can be viewed as meta-parameters). In our empirical evaluation we used $\epsilon=10^{-9}$ for which we achieved the desired performance with $l=70$ in all generated samples. Since we also consider the probability of having an empty system, this corresponds to covering the probability of up to 69 customers in the system. 

\begin{remark}\label{remark:samplingtime}
In term of run-times, sampling a single service time distribution and computing the output takes on average 0.56 seconds. For the sizes of training, validation, and test data sets provided in Table \ref{table:datasets}, sampling the entire data took less than 7 hours using a cluster with 30 nodes. Each node uses Intel Xeon Gold 5115 Tray Processor, with RAM memory of 128 GB. 
\end{remark}


\section{Deep Learning}\label{sec:deep}

In this section, we provide details into our deep learning model including our choice of network architecture and the loss function. We start by describing the input pre-processing step in Section \ref{sec:intput preprocessing} and then the network architecture in Section \ref{sec:network}.

\subsection{Input pre-processing}\label{sec:intput preprocessing}

Once the sampling procedure in Algorithm~\ref{algo:main} is completed we end up with pairs of arrival rate and service time distributions (in the form of $\alpha$ and $S$). Hypothetically, we could enter $\alpha$ and $S$ along with their corresponding arrival rates directly into the model. However, we have made a design choice not to do so as it leads to very sparse and inefficient representations
that can have more than $10,000$ values. Instead, we decided to compute the first $n$ moments of the service time distribution, and use it as an input into the deep learning model together with the sampled arrival rate. 

Note that it is common practice in deep learning to normalize the input so that all input values will be within the same range. We wish to enforce range similarity (1) between samples  and (2) within a sample (i.e., arrival rate and service moments). While scaling all service time distributions so that the expected value is 1, achieves between-sample similarity, as we increase the moments order they increase exponentially, violating (2). Thus we apply a $log$ transform to all service time moments. This resolves (2). Since the first moment in all cases is equal to $1$, we omit it.

\subsection{Network Architecture}\label{sec:network}
\begin{figure}
\centering
\includegraphics[scale=0.5]{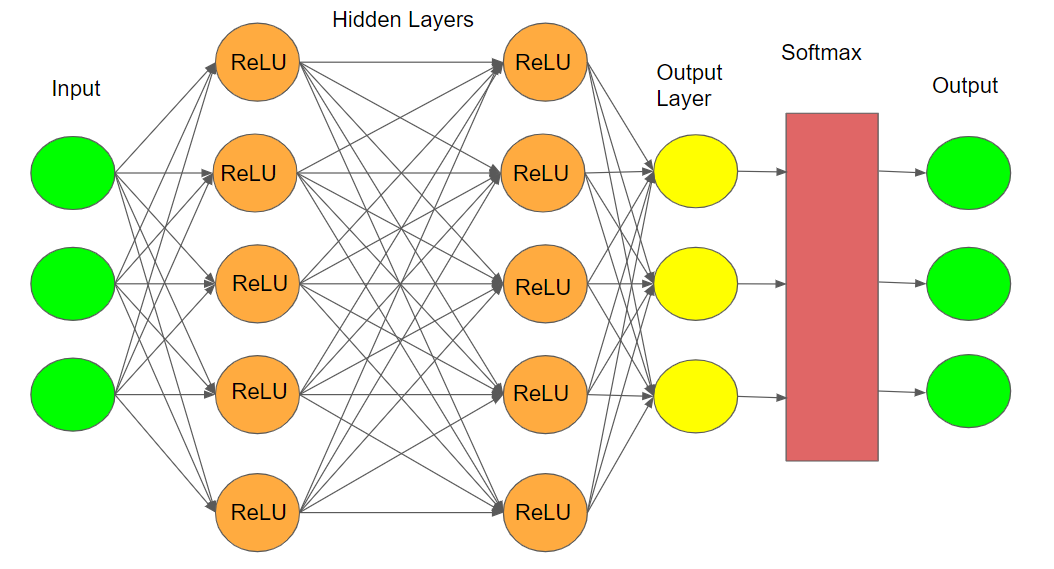}
\caption{Diagram of our neural network. }
\label{fig:NN_diagram}
\end{figure}

We deploy a Fully-Connected (FC) feed-forward neural network as depicted in Figure~\ref{fig:NN_diagram}. While other architecture types, such as Convolution Neural Nets (CNN) and Recurrent Neural Nets (RNN) can certainly be considered, we felt neither one was appropriate for this case: CNN is not well suited to our needs as the input size is only a small vector, while RNN is more suited for time-dependant, rather than stationary, systems.

We used the Rectified Linear Unit
(ReLU)~\cite{DBLP:journals/corr/abs-1803-08375} as the activation function for all hidden layers. 
Furthermore, we applied the Softmax function to guarantee that the output-layer 
weights sum up to 1 (while this is not a common use of the Softmax function, it is quite convenient in our case as our output must sum to 1). An
exhaustive tuning of all the related hyper-parameters is out of the
scope of this paper, and commonly used default values for the
architecture of the neural network were used throughout our experiments. 

\noindent \textbf{Loss function} We now present the loss function for our network. For the training batch size $B$ (a hyper-parameter), let $Y$ and $\hat{Y}$ present a matrix of size $(B, l)$ of the true values of the stationary queue length distribution and the network prediction, respectively. 
Our loss function $Loss(\cdot)$ is given by

\begin{equation}  \label{eq:loss}
     Loss(Y,\hat{Y}) =   \frac{1}{B}\sum_{i=1}^{B}\sum_{j=0}^{l-1}|Y_{i,j}-\hat{Y}_{i,j}|+ 
     \frac{1}{B}\sum_{i=1}^{B}max_j(|Y_{i,j}-
     \hat{Y}_{i,j}|),
\end{equation}

\noindent where the values $Y_{i,j}$ and $\hat{Y}_{i,j}$ refers to the $i^{th}$ sample in the batch and the probability of having $j$ customers in the system for the respective distributions, and, as discussed in the previous section, $l$ is the output length, which is fixed to 70. 
From equation~\eqref{eq:loss} we observe that the loss function is comprised of two terms.  
The first term aims to minimize the overall absolute distances between the true and predicted distributions, while the second term seeks to minimize the maximum distance. This functional form is motivated by typical queueing applications where we prefer a few smaller errors to a a single large one (particularly since large errors tend to occur in the tail, which may be particularly important for providing service level guarantees). While the are many other measures of distance between two distributions, we felt they were 
less suitable for our purposes. For instance, cross-entropy is more suitable for zero-one classification tasks, while KL-divergence is less suitable for distributions having numerous zero values, which is a common scenario in stationary queue-length distributions. 

The loss function refers to a single batch size denoted by $B$, a hyper-parameter that is be tuned during  training. All selected hyper-parameters including $B$ are described in Section~\ref{sec:deep_architecture} below.

\section{Model Tuning and Experimental Evaluation}\label{sec:evaluation}

In this section we describe training and validation of our model, and provide an extensive empirical evaluation of our approach on test data. 


\subsection{Experimental Setting}\label{sec:experimental setting}
In this section we discuss the different data-sets used to train and validate our model, the training procedure, the metrics used to evaluate our model, and the architecture and values of hyperparameters for our final model.

\subsubsection{Data generation}
We used our sampling procedure described in Section \ref{sec:sampling} above to produce three different data sets: the training, validation, and test data sets. Their sizes  
are provided  in Table~\ref{table:datasets}.
\begin{table}[t]
\small
\centering
\begin{tabular}{||c| |c|c|c||} 
 \hline
 Data set & Training & Validation & Test \\ [0.5ex] 
 \hline
 Size & 1,200,000 & 60,000 & 60,000  \\ 

 \hline
\end{tabular}
\caption{Dataset sizes.}
\label{table:datasets}
\vspace{-1em}
\end{table}

\noindent Training the deep learning model (namely, its weights) is performed on the training set.
The validation set is used during training for the purpose of tuning hyper-parameters of the network.   The model performance is assessed based on the test data set.


\subsubsection{Model Training and Tuning}\label{secP:indepVariables}

The main logical flow of our process is as follows. Once all three data-sets are sampled and pre-processed, including the computation of the output, we train our model for every value of the number of input moments  $n = 2,..,20$. For each $n$, we fine-tune our model, i.e., choose the best hyper-parameters (e.g., number of hidden layers, learning-rate schedule, optimizer), based on our validation set. Different hyper-parameters settings are compared via a simple metric, $Metric1$, given by the following equation:
\begin{align}\label{eq:acc}
Metric1(Y,\hat{Y}) =  \frac{1}{N}\sum_{i=1}^{N}\sum_{j=0}^{l-1}|Y_{i,j}-\hat{Y}_{i,j}|,   
\end{align} 
where $Y$ and $\hat{Y}$ correspond to two matrices of size $(N, l)$ of the true values of the stationary queue-length distribution and of the network's predicted values, respectively. Here $N$ is the size of the entire validation data set. Note that $Metric1$ is the absolute sum of errors (i.e., absolute distances between the true and predictied distributions) computed for each sample and then averaged over the entire validation data set. The main advantages of this function is that it has a straightforward probabilistic meaning and it covers the entire PDF of the stationary queue length. Also, for each sample it provides an upper bound on the maximum difference between the two CDF functions, which has statistical justification when examining distances between two distributions, e.g., in the Kolmogorov–Smirnov test.

For each value of $n$ we choose settings that minimizes the value  of $Metric1$. Once the optimal settings for each $n$ are determined, we search for the `best' value of $n$, i.e., the value that achieved the minimal value of $Metric1$ over the validation set.  However, if we reach convergence, i.e., for different values of $n$ the value of $Metric1$ does not change by more than 
 $\epsilon  = 10^{-6}$, we select the smallest value of $n$ for which this convergence is achieved. The intuitive idea is that if adding more moments will give only a negligible added benefit value, then it is not worth doing.  

\subsubsection{Analyzing the number of service time moments}

\label{sec:moms_analysis}

\begin{figure}[t]
\centering
\includegraphics[scale=0.6]{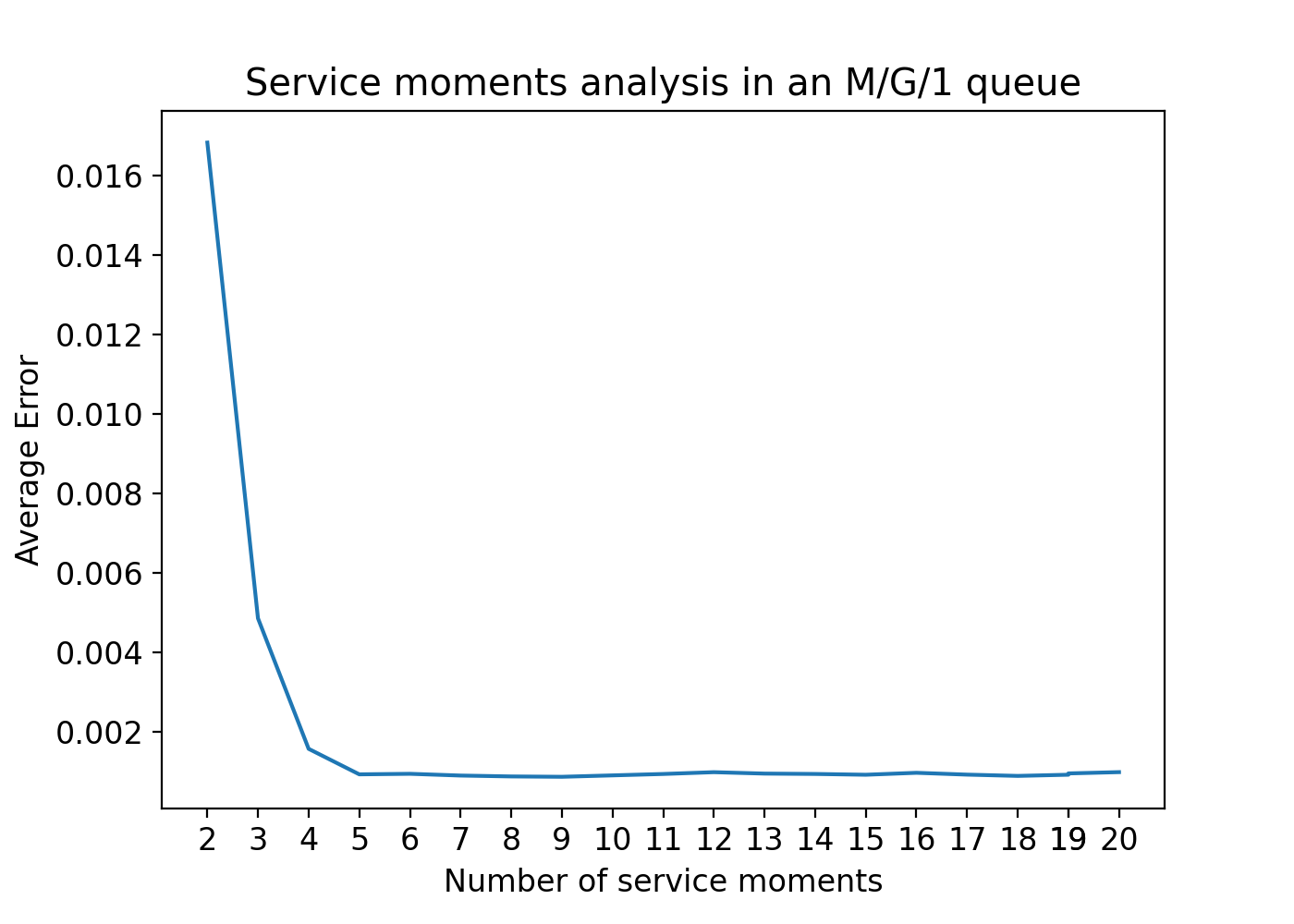}
\caption{Service moments analysis.  }
\label{fig:moments_analysis}
\end{figure}

In Figure~\ref{fig:moments_analysis} we display our results of the service moment analysis. We compute the value of $Metric1$ from Equation~\eqref{eq:acc} as a function of the number of moments. As shown in the figure, convergence is achieved once the number of moments reaches 5. This implies that only the first 5 service time moments appear to determine the dynamics of the $M/G/1$ queue. This result is somewhat surprising as two distributions sharing the same first 5 moments may differ significantly, yet will, apparently, result in the same stationary behavior. On the other hand, as discussed earlier, the average queue length is determined by only the first two service time moments, which makes this result more sensible.  

\subsubsection{Deep learning model architecture for $n=5$ }\label{sec:deep_architecture}

Once the best value of $n$ has beem
determined to be $5$, we fixed the values of all other hyper-parameters.  The resulting network is described below.

We have trained a fully connected neural network described in Section~\ref{sec:deep}. In total, there are 5 hidden layers with  14,399 parameters. We have used the Adam optimization algorithm~\cite{Kingma14} for an iterative
update of the network weights based on the training data. The corresponding weight decay value for avoiding over-fitting was $10^{-5}$. The selected training batch size was $128$. The learning rate decayed exponentially, starting from 0.01. The number of trained epochs was 300. The model was trained on an NVidia A100 GPU. For further details about the chosen deep learning model see Appendix~\ref{sec:appendix1}.

In terms of computational complexity this is considered to be a small network (a large network typically contains several millions parameters\footnote{For example the ResNet101, which is a commonly used architecture in computer vision has 44.5 million parameters.}). This leaves a lot of room for adding network complexity as machine learning methodology is extended to more complex  queueing systems.


\subsection{Results}\label{sec:result}

In this part, we present the evaluation of our machine learning system on the test dataset. Throughout this section we use the ``best'' network with $n=5$, tuned as described above. 

For each sample in the test set we compare the true stationary queue length distribution to its counterpart predicted by our model. In addition to $Metric1$ (absolute distance) given by equation (\ref{eq:acc}) above, where $N$ now represents the  size of the test data set, we also define the following additional metric which focuses on the inverse of the CDF of the number of customers in the system. Specifically, we compute the average relative error for each of the $6$ different percentile values, namely, $25\%, 50\%, 75\%,90\%, 99\%, 99.9\%$, as follows:   
\begin{align}\label{eq:metric2}
Metric2(Y,\hat{Y}, percentile) = \frac{1}{N}\sum_{i=1}^{N} \frac{F_{Y_i}^{-1}(percentile)-F_{\hat{{Y_i}}}^{-1}(percentile)}{F_{Y_i}^{-1}(percentile)}.
\end{align}
This metric is particularly effective in detecting differences in the tails of the two distribution functions; as noted earlier accurate estimates of tail probabilities are particularly important in many practical applications of queuing systems.

\subsubsection{Performance analysis for the ``best'' model}

\begin{figure}
\centering
\includegraphics[scale=0.65]{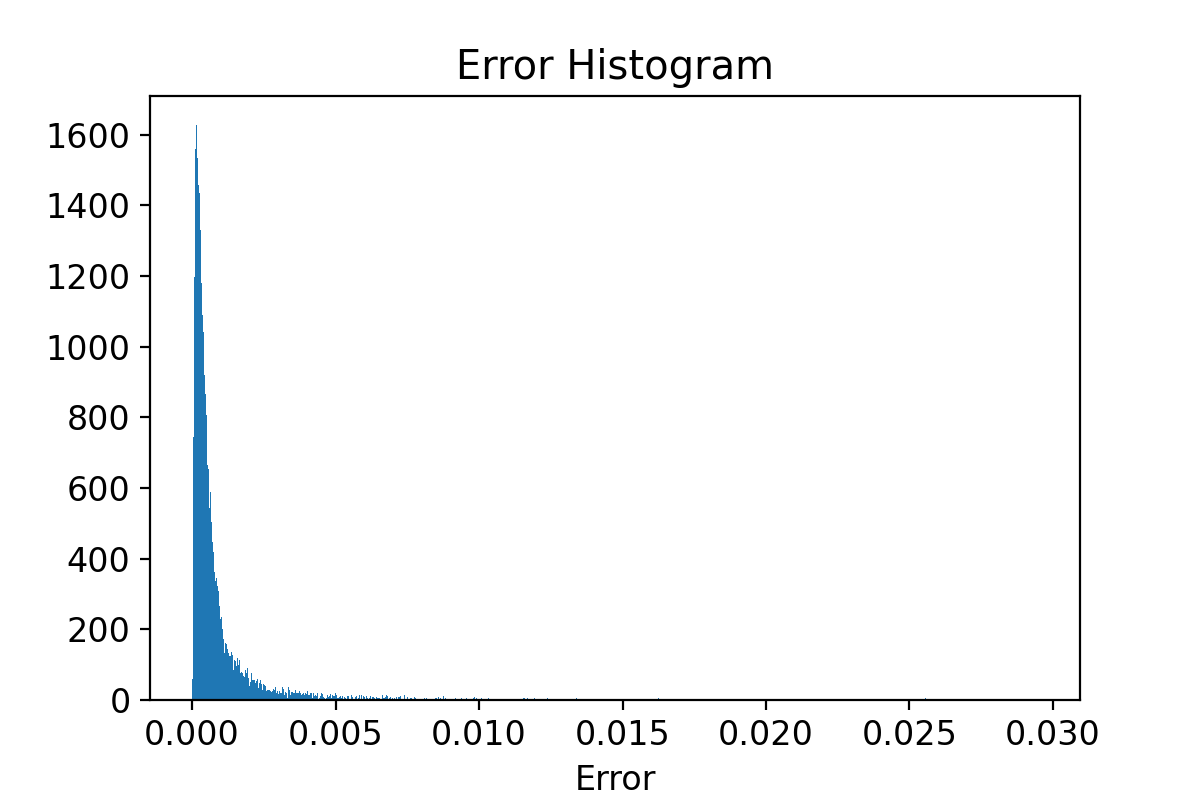}
\caption{An histogram of the error metric presented in Equation~\eqref{eq:acc}.  }
\label{fig:error_hist}
  \vspace{-0.5em}

\end{figure}

In this section we demonstrate how well our deep learning model approximates the stationary probabilities.  The  mean value of the error of our first metric (i.e., Equation~\eqref{eq:acc}) under $n=5$ is $0.0009$. In Figure~\ref{fig:error_hist} we present the error histogram of $Metric1$, to provide a better picture of our performance. In  Table~\ref{table:precetiles_error} we present the performance of our model under both metrics. 

Note that the interpretation of values for $Metric1$ and $Metric2$ are quite different. For $Metric1$ (the seond row of the table) the values represent the exact percentiles of the absolute 
difference between true and predicted distributions over the test data set.  For example, the $99.9$-th percentile was only $.02$, indicating an excellent agreement between the two distributions.  For $Metric2$ (last row of the table) the values represent relative absolute differences between the true and predicted distribution at each percentile level.  For example, 
the largest relative difference was only $0.412\%$ achieved at the $99.9$-th percentile.

\begin{table}[t]
\small
\centering
\begin{tabular}{||c |c|c|c|c|c|c||} 
 \hline
 Percentile & 25\%& 50\%& 75\% & 90\%& 99\%& 99.9 \%  \\ [0.5ex]
 \hline\hline

Metric 1 &  0.000220 & 0.0004272 & 0.00087 & 0.002001  & 0.008757 & 0.02001 \\ [0.5ex] 
 \hline\hline

Metric 2 & 0.000342 & 0.000718 & 0.00094 & 0.000897 & 0.00148 & 0.00412  \\ [0.5ex] \hline

\hline
\end{tabular}
\caption{Deep learning model performance on Test Data.}
\label{table:precetiles_error}
\end{table}

\noindent As demonstrated in Figure~\ref{fig:error_hist} and in Table~\ref{table:precetiles_error} the errors are very small, where in 99\% of the cases the error is smaller than 0.01. This implies that our predictions are extremely reliable.  Furthermore, as the result of $Metric2$ suggest, the predicted distribution produced by our model remains very accurate in the tail as well, where the average error for all percentiles is very low.

\section{Case Study: Deep Learning vs. Analytical Queueing Methods}
\label{sec:casy_study}

The results presented in the preceding section show that our deep learning network is able to achieve excellent predictive accuracy on the test data set. However, the experimental conditions used in the current section could be criticized as not very reflective of ``real-life'' settings: after all, for each test sample the network was supplied with the exact values of the first 5 moments of the service distribution, as well as the exact arrival rate.  In more realistic settings, only an empirical set of data observations for service times and arrivals is likely to be available. How can our network be expected to perform under these conditions?  We aim to provide an answer in the current section.

We shall assume that instead of having a closed-form expressions for the inter-arrival and service time distributions, we are only provided with the  timestamps of arrivals, service commencements, and service completions - essentially the system event log. This is the case in most real-life situations.

From the managerial point of view, given past data one may wish to ask ``what-if?'' questions. For example, what if the arrival rate will change next month, while the service process remains the same? How will it effect system waiting times? In this scenario, we would wish to analyze an $M/G/1$ queue where the arrival rate is given, but the service times are known only via the data sample.  

Given such data, and assuming the system is known to operate as an $M/G/1$ queue, there are two approaches for obtaining the stationary distribution: the "analytic approach" and the "deep learning" one. 
Under the analytical approach, we first use the data to fit some parametric distribution to service times (we will assume a PH distribution is used), and then we apply one of the available analytical expressions to obtain the stationary distributions (we will assume the QBD method is used). 
The machine learning approach (assuming a trained  network is available) is simpler: we only need to compute the first five service time moments from the data, and then we apply the deep learning network to obtain the predicted stationary distribution.

Both methods will incur some errors. In the analytic approach the main accuracy loss can be expected when fitting the service time distribution;  numerical errors from the application of the QBD calculation should be very small (and controllable). In the deep learning method there are two potential sources  of error: due to  the estimation of the service moments from the data, and due to the prediction error of the deep learning model.  

The case-study was executed as follows: 
\begin{enumerate}
    \item We sample a service distribution using our Algorithm 1 from Section \ref{sec:sampling}, scale the mean service time to $1$, set the service rate to $0.85$ and then compute the corresponding stationary queue length distribution.  This gives us the ``Ground Truth".
    \item Next, we sample 50,000 data points from the service time distribution producing our  ``observed'' service time sample.
    \item We apply the analytic approach to the data. This consists of using the EM algorithm to fit the distribution (details are provided below), followed by the QBD method. This gives us the ``Analytic'' estimate.
    \item We apply the deep learning approach producing the ``Deep Learning'' estimate.
\end{enumerate}

\begin{figure}[!tbp]
  \centering
  \begin{minipage}[b]{0.45\textwidth}
    \includegraphics[width=\textwidth]{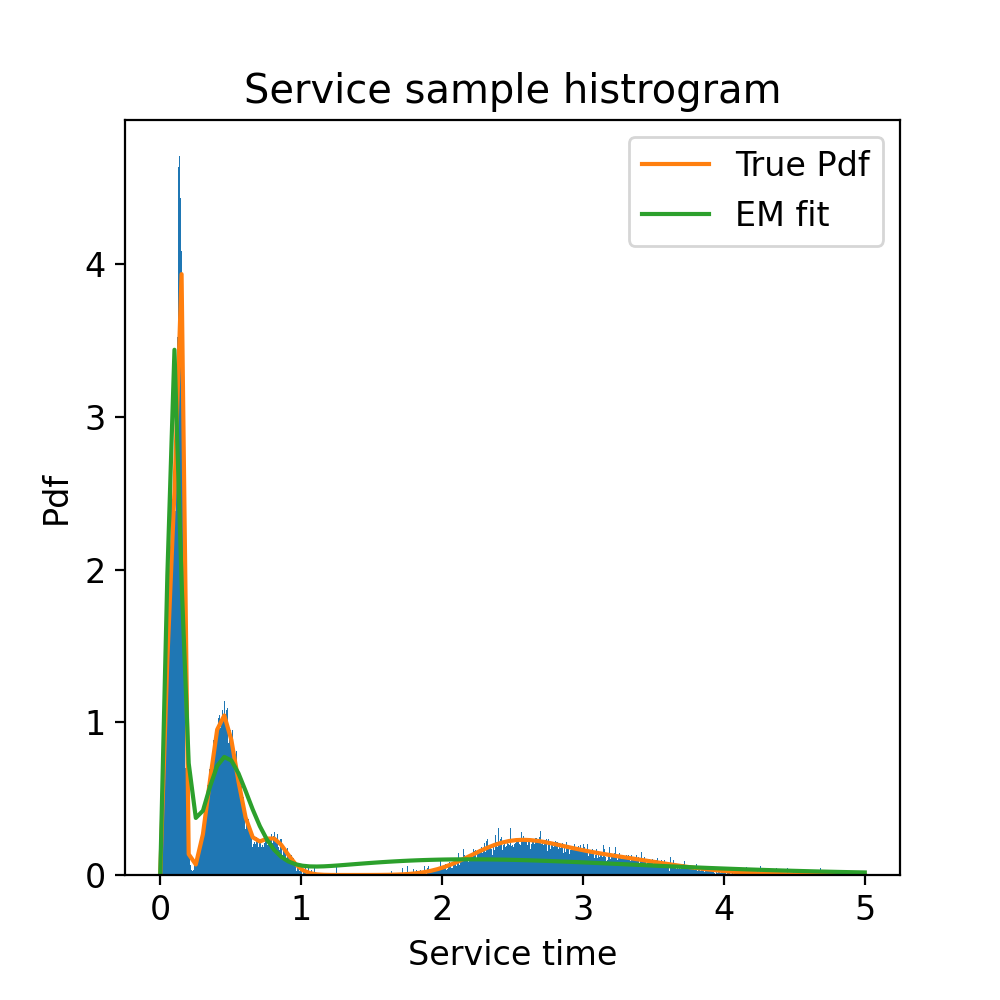}
    \caption{Service data sample}
    \label{fig:service_sample}
  \end{minipage}
  \hfill
  \begin{minipage}[b]{0.45\textwidth}
    \includegraphics[width=\textwidth]{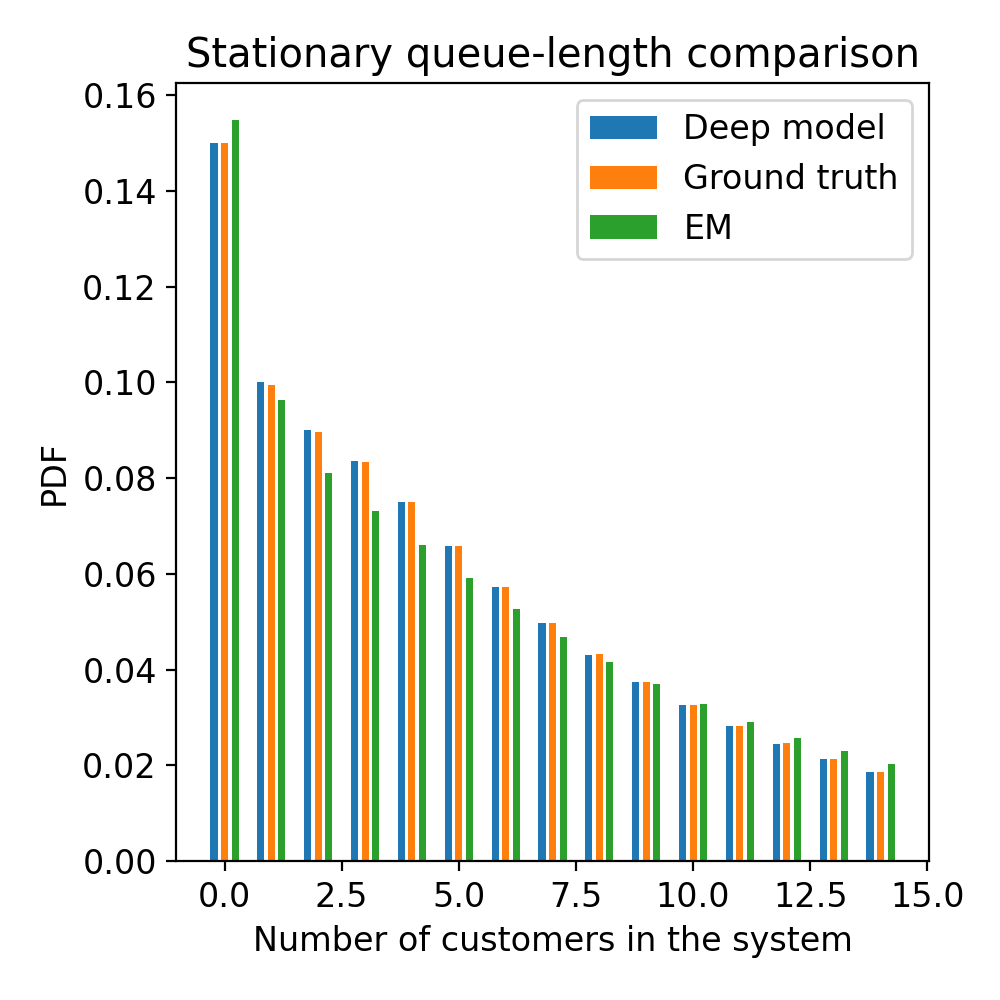}
    \caption{PDF of number of customers in the system.}\label{fig:case_study_pdf}
  \end{minipage}
\end{figure}

Note that while the arrival rate is fixed at $0.85$ (even if arrivals were provided as observational data, the estimation of the arrival rate is straightforward  by using the inverse of the average), fitting a  service time distribution is not easy. The  true and empirical distributions of service times along with a histogram of the data are
presented in Figure~\ref{fig:service_sample}. 

The best practice is to use the EM algorithm to fit the closest PH distribution (see~\cite{Fackrell098}). The weakness of the approach is its computational complexity.  The runtime of the EM algorithm depends on the size of the PH, and thus we get a clear trade-off between accuracy and runtime. In our experience, even for PH of size 20 (this refers to the dimension of the transition matrix) it takes over an hour to run the EM algorithm in most cases\footnote{This depends of course on the number of samples in the data set.}. Moreover,  
the runtime grows exponentially with the size of the PH. In our case-study we used maximum PH size of 20.


For the deep learning approach the first five moments were fitted by employing an unbiased estimator directly from the data. That is, $\bar{X^i} = \sum_{i=1}^{N} \frac{X^i}{N} $, where $X$ is the service time, $N$ is the sample size, and  $i=1,2,3,4,5$.


In Table~\ref{tab:case_study_moms} we present the first $5$ moments of service times estimated by each method (for the analytic approach these were computed from the PH distribution obtained by the EM method). We see that (not surprisingly) the direct estimates used by the Deep Learning approach are far more accurate, particularly with respect to higher-order moments.  This points to substantial differences between the service time distribution estimated by the EM method and the ground truth distribution from which the data was generated. This can also be observed from the comparison of actual and fitted distributions on  Figure~\ref{fig:service_sample}. 

Table~\ref{tab:case_study_moms} also presents pre-processing running times (for the analytic approach this is the running time of the EM method, for the deep learning time this is the time to compute the $5$ moments) and the algorithm running times (QBD computation times for the analytic and ground truth approaches; network computation time for the deep learning approach).  The largest difference is in the very long pre-processing time for the analytic approach compared to near-instantaneous time for the deep learning one\footnote{The runtimes presented in Table~\ref{tab:case_study_moms} corresponding to the `EM' procedure computed on the same server detailed in Remark~\ref{remark:samplingtime}, i.e., Intel Xeon Gold 5115 Tray Processor, with RAM memory of 128 GB.  }. 

\begin{table}[t]
\small
  \centering
    \begin{tabular}{|l|r|r|r|}
    \hline
          & \multicolumn{1}{l|}{Analytic (EM)} & \multicolumn{1}{l|}{Deep Learning} & \multicolumn{1}{l|}{Ground Truth} \\
    \hline
    First moment & 0.994 & 0.994 & 1 \\
    \hline
    Second moment & 2.705 & 2.398 & 2.418 \\
    \hline
    Third moment & 10.093 & 7.037 & 7.106 \\
    \hline
    Fourth momnet & 44.131 & 21.961 & 22.19 \\
    \hline
    Fifth moment & 217.936 & 71.227 & 71.97 \\
    \hline
    Pre-procsseing time (sec) & 4320  & 0.3   & 0 \\
    \hline
    Running time (sec) & 0.16  & 0.12  & 0.44 \\
    \hline
    \end{tabular}%
      \caption{Service moments and runtimes}

  \label{tab:case_study_moms}%
    \vspace{-1.5em}

\end{table}%

Next, we compare the accuracy of the two approaches. 
The histogram of stationary queue-length distributions under the three approaches is presented on  Figure~\ref{fig:case_study_pdf} and the values of $Metric2$ for different percentiles are shown on  Table~\ref{tab:prectiles}. We also computed $Metric1$ for both techniques.

While both the analytical and the deep learning approach do a fairly good job of estimating the stationary queue length distribution, the deep learning approach is the clear winner. The absolute difference between the estimated and the true distributions  according to Equation~\eqref{eq:acc} (i.e., $Metric1$) for the analytic method is 0.094, while it is only 0.002 for the deep learning method. Figure~\ref{fig:case_study_pdf} and   Table~\ref{tab:prectiles} show that the difference in performance is mostly due to the worse fit in the tail of the distribution for the analytic method. While the deep learning model incurs minimal errors at every percentile on Table~\ref{tab:prectiles}, the errors under the analytic approach are quite substantial starting with $75$-th percentile. 

To conclude, this case-study demonstrates the practical capability of our model, especially in answering ``what-if?'' questions. Whereas, moments are easy to estimate from the data, fitting a service time distribution is both computationally burdensome, and can lead to significant accuracy loss.


\section{Discussion \& Limitations}\label{sec:discussion}

\label{sec:limitations}
In this section, 
we discuss some of the main limitations of our approach,  propose several solutions to overcome them, and compare our deep learning model to employing simulation in queueing systems for the task of performance prediction.

\subsection{Limitations of our Approach}
The main challenge 
is extending the deep learning approach to more complex and realistic queuing systems for which, unlike the $M/G/1$ case analyzed above, no analytical solutions are available. The issue is obtaining large and 
reliable data sets on which the deep learning network can be trained. We distinguish between two types of data sets. 

\paragraph{Real data.} Real
data that describes the queueing dynamics of the system can be extracted from databases of many organizations. The advantage of the real data is that ``ground truth'' of actual system performance is directly available.  The main problem is that one cannot typically generalize it to other systems, or even to different time periods for the same system, as the data essentially captures a single realization for a single queuing system. This issue can potentially be addressed by combining different data sets together, and by applying principles from transfer learning. Transfer learning is a machine learning method where a model developed for a task is reused as the starting point for a model on a second task . This is common practice in deep learning, and it is empirically proven to be efficient (see~\cite{10.1007/978-3-030-01424-7_27}).
\begin{table}[t]
\small
  \centering
    \begin{tabular}{|r|r|r|}
    \hline
    \multicolumn{1}{|l|}{Percentile} & \multicolumn{1}{l|}{Analytic} & \multicolumn{1}{l|}{Deep learning}   \\
    \hline
    25    & 0.5     & 0.5     \\
    \hline
    50    & 0     & 0      \\
    \hline
    75    & 0.222    & 0      \\
    \hline
    90    & 0.125   & 0    \\
    \hline
    99    & 0.187    & 0    \\
    \hline
    99.9  & 0.163    & 0.020     \\
    \hline
    \end{tabular}%
      \caption{ Comparison of $Metric2$ }

  \label{tab:prectiles}%
  \vspace{-2.5em}
\end{table}%
\paragraph{Synthetic queueing data.} Such data is sampled from a queueing system with a known structure, as we have done in this paper. However, producing training output is not straightforward since we are typically interested in applying machine-learning models to systems for which closed-form solutions do not exist. This situation imposes a catch: we can use our method to learn ``queueing theory'', but only for queueing models that the theory already knows how to solve. 

We discuss several approaches for dealing with this issue. The most direct extension is using our PH sampling algorithm to produce synthetic data for systems where arrival and/or service processes are of PH form (i.e., $PH/M/1$, $PH/PH/1$ systems) - such systems often have analytical solutions. As our results indicate, PH-type distribution is a good proxy for the general case, thus extensions to, for example, $G/G/1$ queue should be relatively easy.  

For more complex systems, we may be able to perform transfer learning from tractable queueing systems. An example for such an approach would be 
the task of deriving stationary queue-length distribution of a $G/G/c$ queue. We can use data from $M/M/c$ and $PH/PH/1$ queues, which both have analytical solutions.   

Another option is to use simulation models to generate training samples.  In principle, a simulation model can deal with very high level of complexity of the underlying system and produce stationary distributions. The practical limitation is that the runtime until convergence tends to be long, even for relatively simple systems (particularly in the multi-server case). Thus, while we were able to generate over a million training samples in a matter of hours using our sampling algorithm, this may not be possible with simulation models.

While constructing a large training data set with simulation alone is not practical,
it may be possible to train a simpler model (e.g., $G/G/1$ with distinct customer classes) and then (again) apply 'transfer learning' with fewer data points on a more complex model (e.g., the $G/G/c$ counterpart).

\subsection{Discussion: Machine Learning vs. Simulation}

Simulation is often the tool of choice
when the underlying system is intractable, i.e., a closed-form solution of the steady-state distribution is unavailable. If the model 
is correct, we are guaranteed that if we simulate for long enough our results will be very accurate. However, there are several advantages we would like to note when using our approach compared to simulation. The first is runtime: as noted earlier, it takes a long time for simulations to converge, while a deep learning model provide predictions immediately after it is trained. The second benefit of our approach lies in the deep learning model itself. Although these models
are often considered  black-boxes system abstractions, similarly to simulation, they may provide insights beyond what can be achieved with simulation models. For example, the moments analysis we performed done in Section~\ref{sec:moms_analysis} would be impossible when using simulation, as the latter require a full specification of the service time distribution. 

More generally, the ability of the deep learning model to work with limited data (first $5$ moments in our case) rather than a full specification of the distribution gives it potentially great advantages over the simulation approach.

\subsection{Deep Learning for Fitting Distributions to Data}

Before leaving this section we should mention one more interesting application of our methodology.  Fitting a distribution to data is of general interest in many contexts, both within and beyond queuing applications. Although the PH distribution family is very versatile,  fitting it to data is not a simple task. As we saw in Section \ref{sec:casy_study}, the standard approach of using EM algorithm is neither accurate, nor computationally efficient. 

In our model we used moments to represent a given PH distribution. However, we could replace the roles, using moments computed from the data as inputs, while the PH representation (i.e., the initial state distribution vector and the transition matrix) would become the output. Our sampling algorithm can easily generate training samples, and - as we saw in Section~\ref{sec:casy_study} - computing moments directly from the observational data is both accurate and efficient. 

In fact, there are several studies that deal with fitting PH given the first $n$ moments. For example in~\cite{doi:10.1081/STM-200056210}, the authors present an explicit method to compose minimal
order continuous-time acyclic phase type (APH) distributions given only the first three moments. Yet their method does not always find a solution and it is limited to the first three moments. We feel our deep learning approach can be re-purposed to provide more efficient solutions to this important problem. 

\section{Conclusion}\label{sec:conclusions}

In this paper we empirically demonstrated the ability of machine to analyze a general queueing model, namely the stationary queue-length distribution of an $M/G/1$ queue. Our results show that machines can accurately learn these distributions using only a relatively compact deep learning architecture. This 
gives hope for future studies where more complex queueing models (and thus more complex deep learning models) will be incorporated.

The unique method of evaluating the stationary queue length distribution using only service time moments instead of the full analytical representation, allows us to (empirically) evaluate the number of service time moments required for determining the stationary queue-length distribution. Surprisingly, the accuracy does not significantly improve when we go beyond the $5^{th}$ moment.  We also provided a case study that describes a real-life setting when data samples are available. We demonstrated that the deep learning model can outperform analytical (data-driven) methods both in terms of accuracy and runtime. 

Finally, we believe that this approach will be very useful in future studies, as our deep learning approach excelled in capturing the stochastic dynamics of a stationary queueing model. As discussed in the preceding section, extensions to $G/M/1$, $G/G/1$ and similar models should be rather direct. Applications to more complex settings, as well as to non-queuing settings such as fitting PH distribution to data, are also promising.


\bibliographystyle{plain}
\bibliography{main}

\newpage
\begin{appendix}

\section{Deep Learning model}

\begin{figure}[h]%
    \centering
    \subfloat[\centering Training loss]{{\includegraphics[width=6cm]{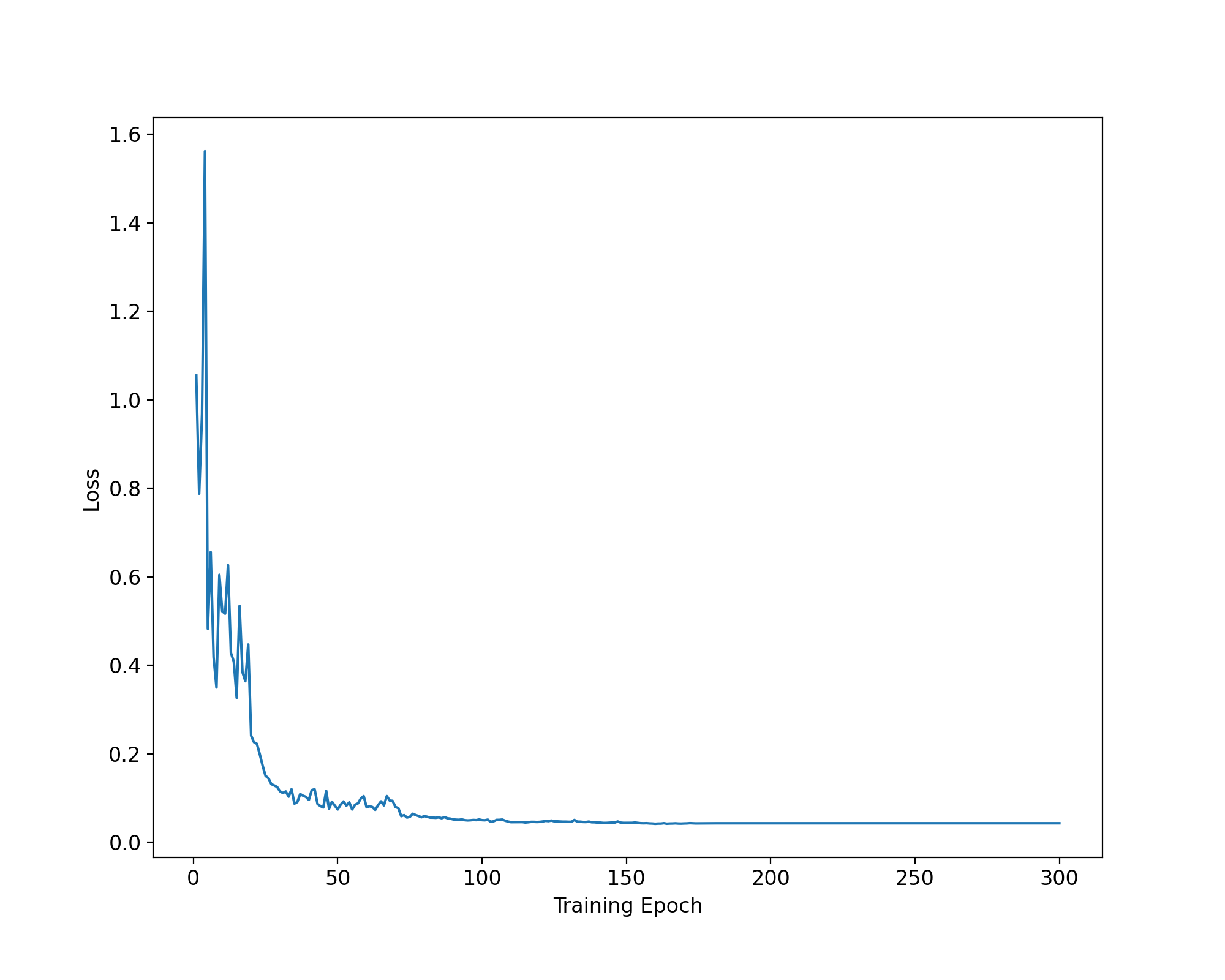} }}%
    \qquad
    \subfloat[\centering Metric1  ]{{\includegraphics[width=6cm]{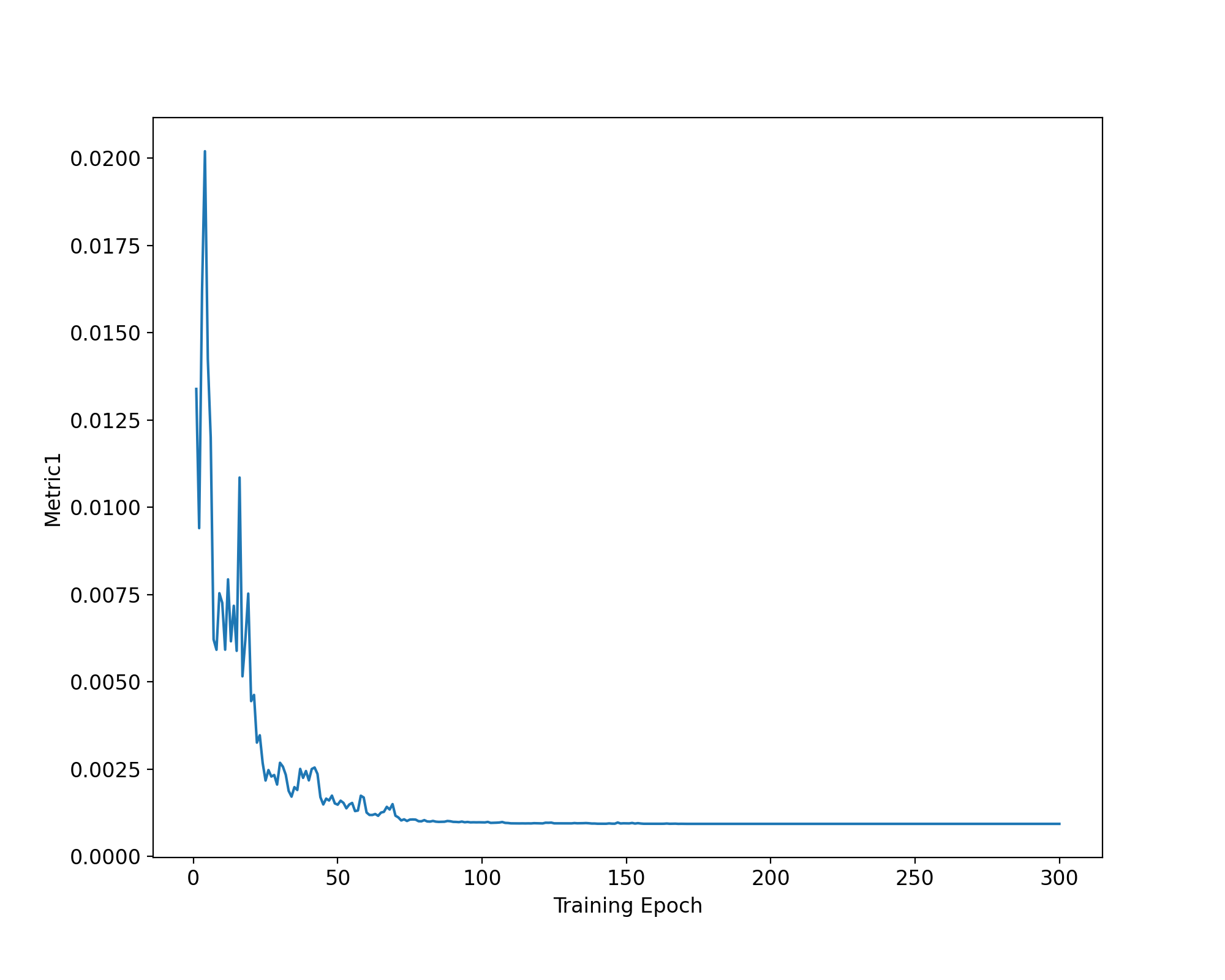} }}%
    \caption{Training Phase}%
    \label{fig:deep_epochs}%
\end{figure}

In this section we provide additional details on the chosen deep learning model. In Figure~\ref{fig:deep_epochs}, we present the training loss and the value of $Metric1$ on the validation set as a function of the training epoch. In Table~\ref{table:deep_nodes} we specify the network specific architecture.  

\begin{table}[h]
\centering
\begin{tabular}{||c |c|c|c|c|c||} 
 \hline
 Hidden layer & 1& 2& 3 & 4& 5  \\ [0.5ex]
 \hline\hline

Nodes &  30 & 40 & 50 & 60  & 60 \\ [0.5ex] 
 \hline\hline
\end{tabular}
\caption{Network nodes.}
\label{table:deep_nodes}
\end{table}

\label{sec:appendix1}

\end{appendix}

\end{document}